\pdfoutput=1

\documentclass[11pt]{article}

\usepackage[preprint]{coling}

\usepackage{times}
\usepackage{latexsym}

\usepackage[T1]{fontenc}

\usepackage[utf8]{inputenc}

\usepackage{microtype}
\usepackage{url} %
\usepackage{booktabs} %
\usepackage{amsfonts} %
\usepackage{nicefrac} %
\usepackage[caption=false]{subfig}
\usepackage{amsmath}
\usepackage{bm, upgreek}
\usepackage{amssymb}
\usepackage{array, booktabs}
\usepackage{multirow}
\usepackage{mathtools}
\usepackage{arydshln}
\usepackage{tabularx}
\usepackage{color}
\usepackage{soul}
\usepackage{lipsum}
\usepackage{amsthm}
\usepackage{caption}
\usepackage{xcolor}
\usepackage{pifont}%
\usepackage{comment}
\usepackage{cleveref}
\usepackage{enumitem}
\usepackage{diagbox}
\usepackage[most]{tcolorbox}
\usepackage[ruled,vlined]{algorithm2e}
\newcommand{\cmark}{\ding{51}}%
\newcommand{\xmark}{\ding{55}}%
\usepackage{inconsolata}

\newcommand{\Ours}{Intermediate Steps for Lists}
\newcommand{\ours}{\textsc{ISL}}

\newcommand{\ourdata}{\textsc{List2QA}}

\definecolor{readablered}{RGB}{170,0,0}

\title{Structured List-Grounded Question Answering}

\author{
    Mujeen Sung$^{1}$\Thanks{~Work performed during an internship at AWS AI Labs.}\quad Song Feng$^{2}$\quad James Gung$^{2}$\quad \\\bf{Raphael Shu$^{2}$ \quad Yi Zhang$^{2}$\quad Saab Mansour$^{2}$} \\
    Kyung Hee University$^{1}$\quad AWS AI Labs$^{2}$\\
    \texttt{mujeensung@khu.ac.kr} \\
    \texttt{\{sofeng, gungj, zhongzhu, yizhngn, saabm\}@amazon.com}
}

\begin{document}
\maketitle
\begin{abstract}
Document-grounded dialogue systems aim to answer user queries by leveraging external information.
Previous studies have mainly focused on handling free-form documents, often overlooking structured data such as lists, which can represent a range of nuanced semantic relations.
Motivated by the observation that even advanced language models like GPT-3.5 often miss semantic cues from lists, this paper aims to enhance question answering (QA) systems for better interpretation and use of structured lists.
To this end, we introduce the \ourdata{} dataset, a novel benchmark to evaluate the ability of QA systems to respond effectively using list information.
This dataset is created from unlabeled customer service documents using language models and model-based filtering processes to enhance data quality, and can be used to fine-tune and evaluate QA models.
Apart from directly generating responses through fine-tuned models, we further explore the explicit use of \Ours{} (\ours{}), aligning list items with user backgrounds to better reflect how humans interpret list items before generating responses.
Our experimental results demonstrate that models trained on \ourdata{} with our \ours{} approach outperform baselines across various metrics. 
Specifically, our fine-tuned Flan-T5-XL model shows increases of 3.1\% in ROUGE-L, 4.6\% in correctness, 4.5\% in faithfulness, and 20.6\% in completeness compared to models without applying filtering and the proposed \ours{} method. 

\end{abstract}

\section{Introduction}
\begin{figure*}[t]
\centering
\includegraphics[width=0.95\textwidth]{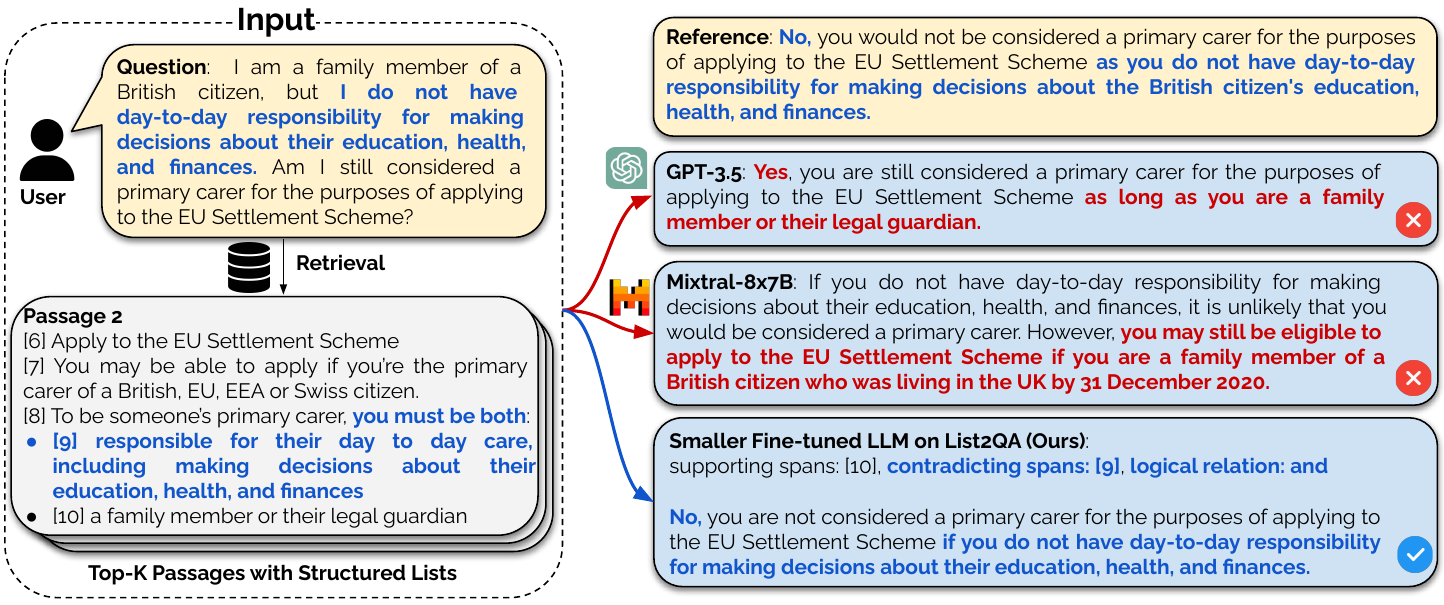}
\caption{
An example of system responses to the user question grounding over list-based content. \textcolor{blue}{Blue} texts indicate semantic cues for correct reasoning, while \textcolor{red}{red} texts indicate incorrect reasoning.
}
\label{fig:overview}
\vspace{-0.3cm}
\end{figure*}

Document-grounded dialogue systems aim to assist users in interactively seeking information from external documents to address various real-world problems with more complex scenarios as seen in customer support. 
While previous work has primarily treated these external documents as unstructured text \citep{campos2020doqa, Feng2020Doc2DialAG, Feng2021MultiDoc2DialMD,
wu2021dialki,
gao2022unigdd, zhao2023causal}, a significant portion of real-world content is presented in structured formats like lists. 
For instance, approximately 45\% of passages in public policies in the UK \footnote{\url{https://www.gov.uk}} comprise lists to effectively present conditions to be verified, action-based steps, or general itemized information (see examples in \Cref{tab:passage_examples}). 
Despite this prevalence, existing research has largely overlooked the nuanced challenges posed in understanding structured list data in relation to the complex background context of users accessing this information. 
Some studies have explored list information for condition verification purposes \citep{Sun2021ConditionalQAAC}, but not in the realistic setup where retrieval is required to differentiate between conditional and non-conditional lists based on user backgrounds. Surprisingly, SOTA models such as GPT-3.5 \citep{chatgpt} and Mixtral-8x7B \citep{jiang2024mixtral} show unsatisfactory performance on nuanced list information, as illustrated in \Cref{fig:overview}, despite their strong results on natural language inference (NLI) and reasoning tasks \citep{qin2023chatgpt, liu2023evaluating, guo2023evaluating}. 

\begin{table}[h!]
\begin{center}
\centering
\resizebox{\columnwidth}{!}{%
\begin{tabular}{@{\extracolsep{4pt}}ll}
\toprule
\multicolumn{1}{c}{\textbf{\textit{A Passage with a Step List}}} \\
\midrule
\begin{tabular}[c]{p{8cm}}
Title: Provide driving tests for your employees \\
Qualifying as a delegated driving examiner \\
Your employees must then:\\
$\bullet{}$ complete an initial training course\\
$\bullet{}$ reach an appropriate standard in the delegated driving examiner theory and practical tests\\
\end{tabular} \\
\midrule
\multicolumn{1}{c}{\textbf{\textit{A Passage with an Option List}}} \\
\midrule
\begin{tabular}[c]{p{8cm}}
Title: Workplace pensions\\
You can get free, impartial information about your workplace pension options from:\\
$\bullet{}$ the Money Advice Service \\
$\bullet{}$ the Pensions Advisory Service \\
$\bullet{}$ Pension Wise if you're in a defined contribution pension scheme \\
\end{tabular} \\
\midrule
\multicolumn{1}{c}{\textbf{\textit{A Passage with a Non-Action Info List}}} \\
\midrule
\begin{tabular}[c]{p{8cm}}
Title: Money and property when you divorce \\
A mediator can help you and your ex-partner agree on how to split money and property.
Mediation is not relationship counselling.
It can help you agree on how you'll divide your assets, including: \\
$\bullet{}$ pensions \\
$\bullet{}$ property \\
$\bullet{}$ savings \\
\end{tabular} \\
\bottomrule
\end{tabular}
}
\end{center}
\caption{
Examples of passages with lists as steps, options, and non-action itemized information.
}
\label{tab:passage_examples}
\vspace{-0.3cm}
\end{table}

Our work aims to address these limitations while testing LLM capabilities for more nuanced list-based content. 
While several benchmarks exist for evaluating QA models on tabular data \citep{chen2021finqa, nan2022fetaqa, oses-grijalba-etal-2024-question}, much less attention has been given to assessing models' abilities on structured lists.
Thus, we propose a novel benchmark called \ourdata{}, designed to evaluate question answering (QA) systems on understanding list semantics with respect to user background. The dataset introduces diverse styles of list content for document grounding, such as specifying logical conditions for validation, describing actionable steps, or simply separating items without explicit logical relations. For QA samples, we construct scenarios where the user background information may align with, contradict, or not address specific list items, which are oftentimes used to determine system responses.

Additionally, we explore pipeline approaches that focus on fine-tuning smaller, more efficient LLMs, demonstrating their potential to outperform larger LLMs on our benchmark dataset. Inspired by recent successes in automated data creation \citep{he2023annollm, choi2024gpts, oh2023ktrl+}, we employ large language models to simulate user queries and system answers grounding over structured lists, and also investigate how to filter low-quality data to further improve performance.

Given that LLMs can often overlook logical relations among list items and their semantic alignment with user-to-item status (See \Cref{fig:overview}), we further investigate whether we can emphasize the semantic cues in lists and improve end-to-end performance. 
Thus, we introduce `Intermediate Steps for Lists (\ours{})', aligning better with how humans interpret list items before responding. 
By explicitly modeling structured list data and user contexts with \ours{}, our method outperforms baseline LLMs on the \ourdata{} dataset.
Specifically, our \ours{} fine-tuned Flan-T5-XL model \citep{Chung2022ScalingIL} shows increases of 3.1\% in ROUGE-L, 4.6\% in correctness, 4.5\% in faithfulness, and 20.6\% in completeness compared to baseline fine-tuning.

\begin{figure*}[t]
\centering
\includegraphics[width=0.95\textwidth]{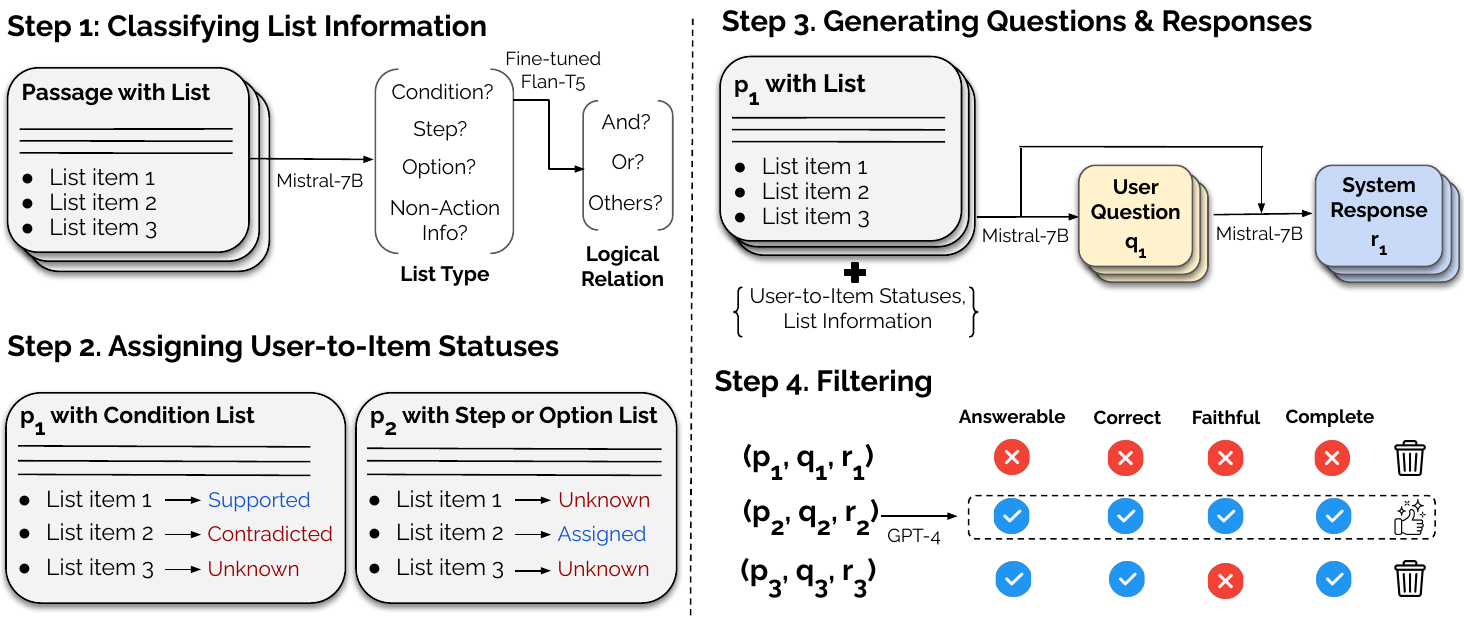}
\caption{
Overview of the \ourdata{} dataset creation pipeline: (Step 1) classifying list types and logical relations in passages with lists, (Step 2) assigning user-to-item statuses for each list item, (Step 3) generating user questions and system responses from the previous steps, and (Step 4) filtering out noisy samples based on four metrics.
}
\label{fig:datacreation}
\end{figure*} 

Our contributions are summarized as follows: 
\begin{itemize}
    \item We present \ourdata{}, a novel benchmark designed to evaluate question answering systems on nuanced, list-based content.
    \item We propose the \Ours{} (\ours{}) method, which enhances alignment with human interpretation of list items before generating responses.
    \item We demonstrate that smaller fine-tuned models using our \ours{} method significantly outperform larger LLMs on the \ourdata{} dataset, setting a new state-of-the-art for this task.
\end{itemize}

\section{\ourdata{}}

In this section, we first formulate the problem of generating system responses based on different types of lists and user scenarios. We then detail the methodology used to create our \ourdata{} dataset, which involves an automated and pipeline-based simulation process utilizing language models. Lastly, we present more details about the dataset for training and validation.

\subsection{Problem Formulation}
To formulate the problem, we first examine the various ways lists can be structured and how user scenarios interact with these lists. Our goal is to develop QA systems capable of providing responses specific to user scenarios based on relevant structured lists.
We categorize lists commonly found in support documents \citep{Feng2020Doc2DialAG, Sun2021ConditionalQAAC} into the following types: conditions for eligibility (`condition'), step-by-step instructions (`step'), options for users to choose from (`option'), and the rest being mostly non-action information without explicit logical relation (`non-action info').

We define our task with an input consisting of a set of passages $\mathcal{P} \in {p_1, \dots, p_N}$, a user question $q_i$, and the system response $r_i$ as the model output. Each passage $p_n$ contains list items. Each user question $q_i$ includes user scenarios or user-to-item statuses.
We follow a retrieval-augmented generation \citep{lewis2020retrieval} pipeline where we employ a passage retriever to select the top-$K$ most relevant passages for each user question.
Based on retrieved passages and a user question, a language model generates system response by reasoning over the information described in the relevant text. Specially, we introduce `\Ours{} (\ours{})', including: 
(1) identifying relevant passages among top-$K$ passages, 
(2) classifying list types (i.e., conditions, steps, items, non-action info), (3) identifying logical relations between condition list items (i.e., and, or), and (4) determining entailment based on user-to-item statuses for condition list items (e.g., supported, contradicted, unknown) or selecting list items aligned with user-to-item statuses for step/option list items.
The final system response, $r_i$, is generated as free-form text based on these intermediate steps.

\begin{table*}[t]
\begin{center}
\centering
\resizebox{\textwidth}{!}{%
\begin{tabular}{@{\extracolsep{4pt}}cll}
\toprule
\textbf{List Types} & \textbf{\begin{tabular}[c]{l}User Questions\end{tabular}} & \textbf{\begin{tabular}[c]{l}System Responses\end{tabular}} \\
\midrule
 \multirow{1}{*}{Condition} & \begin{tabular}[c]{p{7cm}}If there is no break clause mentioned in the tenancy agreement, \textbf{can I as a tenant end the tenancy early with your approval?}\end{tabular} & \begin{tabular}[c]{p{8cm}}Yes, you may be able to end the tenancy early with your landlord's approval, even if there is no break clause in the tenancy agreement.\end{tabular} \\
\midrule
Step & \begin{tabular}[c]{p{7cm}}I'm setting up a business partnership. First, I need to decide on a name for the partnership. \textbf{What should I do next?}\end{tabular} &  \begin{tabular}[c]{p{8cm}}Next, you need to choose a `nominated partner' and register with HM Revenue and Customs (HMRC) to complete the setup of your business partnership.\end{tabular} \\
\midrule
Option & \begin{tabular}[c]{p{7cm}}
I need a private firm to conduct my drivers' medical exam for a lorry or bus license application. \textbf{Are there any other similar services that I can consider instead?}
\end{tabular} & \begin{tabular}[c]{p{8cm}}
You could also consider visiting your GP to complete the medical examination section on the D4 form for your lorry or bus driver license application.
\end{tabular} \\
\midrule
Non-Action Info & \begin{tabular}[c]{p{7cm}}
I'm considering suspending or leaving my course. \textbf{Which types of student finance do I need to stop paying?}\end{tabular} &  \begin{tabular}[c]{p{8cm}}
You need to stop paying your student finance payments for Maintenance Loans, Tuition Fee Loans, and any grants or bursaries you may be receiving.
\end{tabular} \\
\bottomrule
\end{tabular}
}
\end{center}
\caption{
Examples of generated questions and responses for each list type in the \ourdata{} dataset.
}
\label{tab:example}
\vspace{-0.3cm}
\end{table*}

\subsection{Document Corpus}\label{sec:corpus}
We consider two corpus sources ConditionalQA \citep{Sun2021ConditionalQAAC} and MultiDoc2Dial \citep{Feng2021MultiDoc2DialMD}, both of which contains numerous diverse list items in their passages due to the nature of customer support documents. This allows us to develop our own dataset based on their document content. Instead of utilizing their annotations, which focus on plain text information, we use only the unlabeled documents and create all new instances as specified in \Cref{sec:dataset_creation}. 

\noindent ConditionalQA \citep{Sun2021ConditionalQAAC} is a dataset designed for conditional reading comprehension tasks. It uses documents related to public welfare in the UK (e.g., "Apply for Visitor Visa") and includes annotations for yes/no or extractive questions. We use document content from ConditionalQA for creating training, validation, and test sets.

\noindent MultiDoc2Dial 
\citep{Feng2021MultiDoc2DialMD} is based on public support documents such as `ssa.gov' and `va.gov'. We use the unlabeled documents from the MultiDoc2Dial corpus to increase the number of test samples in \ourdata{}. We can further evaluate models on domains that were not seen during training using data samples created from this source.

\subsection{Dataset Creation Pipeline}\label{sec:dataset_creation}

Given the lack of existing datasets designed for building QA systems focused on structured lists, we propose a dataset creation pipeline specifically for addressing this challenge.
First, we extract passages containing lists from unlabeled documents described in \Cref{sec:corpus}.
\footnote{The corpus we use is in HTML format, so we employ the <h> tag as a passage splitter and the <li> tag to indicate passages that contains list items.}
We aim to automate the process based on the advances of LLMs with the pipeline illustrated in \Cref{fig:datacreation}.

\paragraph{Step 1. Classifying list information}
To generate user queries with different contexts, we first identify the type of list information in each passage. Passages are categorized into one of four list types: `condition', `step', `option' or `non-action info'. For passages under the condition type, we also classify their logical relations: `and' or `or'.
To this end, we fine-tuned Flan-T5-XL \citep{Chung2022ScalingIL} using 72 manually annotated training samples.
This approach significantly improved performance, achieving an F1 score of 78.0\% on 30 manually annotated validation samples. See the details of the logical relation classifier in \Cref{abs:logical_relation_classifier}.

\paragraph{Step 2. Assigning user-to-item statuses}

Next, we assign user-to-item statuses to one or more list items. For condition lists, we determine whether each item supports, contradicts, or is unknown in the user scenario. For step and option lists, we randomly select an item and assign it as the user status. For `non-action info' lists, we create questions without specific user background. For condition lists, we aim to provide concluded answers (`yes', `no', `uncertain') that can be derived by considering both the logical relations and the user-to-item statuses of list items. As illustrated in \Cref{tab:condition_status}, if a list has an `And (Conjunctive)' relation where item 1 supports and item 2 contradicts the corresponding user-to-item statuses, the deduced answer is `no'.

\begin{table}[t]
\begin{center}
\centering
\resizebox{1\columnwidth}{!}{%
\begin{tabular}{@{\extracolsep{4pt}}lccc}
\toprule
\textbf{Logical Relation} & \textbf{User-Item Status 1} & \textbf{User-Item Status 2} & \textbf{Short Answer} \\
\midrule
\multirow{3}{*}{And (Conjunctive)} & Supported & Supported & Yes \\
 & Supported & Contradicted & No \\
 & Supported & Unknown & Uncertain \\
 \midrule
\multirow{3}{*}{Or (Disjunctive)} & Supported & Contradicted & Yes \\
 & Contradicted & Contradicted & No \\
 & Contradicted & Unknown & Uncertain \\
\bottomrule
\end{tabular}
}
\end{center}
\caption{
Concluded answers derived from the logical relations and the user-to-item statuses of list items.
}
\label{tab:condition_status}
\vspace{-0.3cm}
\end{table}

\paragraph{Step 3. Generating user questions and system responses}

Based on list types and assigned user-to-item statuses, we sequentially generate user questions encompassing specific user scenarios and system responses. 
For this process, we employ Mistral-7B-Instruct, using three-shot in-context examples for each list type.

\begin{table}[t]
\begin{center}
\centering
\resizebox{\columnwidth}{!}{%
\begin{tabular}{@{\extracolsep{4pt}}lcccc|c}
\toprule
\textbf{Split} & \textbf{Condition} & \textbf{Step} & \textbf{Option} & \textbf{Non-Action Info} & \textbf{Total} \\
\midrule
Train & 524 & 224 & 270 & 369 & 1,387 \\
Dev & 58 & 43 & 36 & 51 & 188 \\
Test & 346 & 161 & 215 & 201 & 923 \\
  \bottomrule
\end{tabular}
}
\end{center}
\caption{
Data statistics of \ourdata{}.
}
\label{tab:data}
\end{table}

\begin{table}[t]
\begin{center}
\centering
\resizebox{\columnwidth}{!}{%
\begin{tabular}{@{\extracolsep{4pt}}ll}
\toprule
\multicolumn{1}{c}{\textbf{\textit{Input}}} \\
\midrule
\begin{tabular}[c]{p{9.5cm}}Given the passages, generate the system response  to the user's question, including intermediate steps:\\
\textbf{Passage 1} \\ {[}1{]} Master's Loan \\ {[}2{]} Healthcare and social work\\ {[}3{]} You can't get a Postgraduate Master's Loan if:\\ $\bullet$ {[}4{]}  you are eligible for an NHS bursary\\ $\bullet$ {[}5{]}  you get a Social Work Bursary\\
\textbf{Passage 2}\\ {[}6{]} Social work bursaries \\ {[}7{]} Eligibility\\ {[}8{]} Social work bursaries are available to eligible social work students who: \\  $\bullet$ {[}9{]} don’t get funding from their employer\\ $\bullet$ \textcolor{blue}{{[}10{]} don’t already have a higher education social work qualification}\\ 
\textbf{Passage 3}\\ {[}11{]} Social work bursaries \\{[}12{]} If you’re training for social work you may get a bursary.\\ {[}13{]} Social work bursaries:\\ $\bullet$ {[}14{]} help with living costs and tuition fees\\ $\bullet$ {[}15{]} don’t depend on your household income\\ \\
\textbf{User question}: \textcolor{blue}{I hold a higher education social work qualification.} Am I eligible for a social work bursary?
\end{tabular} \\
\midrule
\multicolumn{1}{c}{\textbf{\textit{Output}}} \\
\midrule
\begin{tabular}[c]{p{9.5cm}}\textbf{Intermediate Steps:}\\
\quad\textbf{Relevant Passage}: 2\\ \quad\textbf{List Type}: Condition\\ \quad\textbf{User-to-Item Status}: {[}7{]}Unknown, \textcolor{blue}{{[}8{]}Contradicted}\\ \quad\textbf{Logical Relation}: \textcolor{blue}{And}\\\\
\textbf{Response}: No, you are not eligible for a social work bursary \textcolor{blue}{because you hold a higher education social work qualification.}
\end{tabular} \\
\bottomrule
\end{tabular}
}
\end{center}
\caption{
A sample for response generation with intermediate steps. Text in \textcolor{blue}{blue} highlights rationale information for validation.
}
\label{tab:prompt_example}
\end{table}

\paragraph{Step 4. Filtering samples}
We improve the data quality by filtering out instances with inaccurate information or hallucinations. 
To validate the model-based filtering, we have three human annotators label 100 examples across four dimensions: question answerability, and response correctness, faithfulness, and completeness.
We specify that samples are filtered out if questions are unanswerable or responses are incorrect, unfaithful, or incomplete.
We then use GPT-4 \citep{OpenAI2023GPT4TR} to evaluate the same examples and measure the inter-annotation agreement (IAA) between human and model-based filtering, yielding a Cohen's kappa score of 56.1 (see the prompt for verification in \Cref{app:prompt_for_data_verification}).
Finally, we apply model-based filtering to the entire dataset, retaining approximately 51.0\% of the original samples.
This results in 1.4K, 0.2K, and 0.9K samples for training, development, and test sets, respectively.
We use MTLD \citep{mccarthy2010mtld} to assess the diversity of generated samples, achieving scores of 69.9 for questions and 64.4 for answers.
In contrast, the original ConditionalQA dataset scores 26.6 for questions and 10.1 for answers, demonstrating significantly higher diversity in our samples.
Examples of final samples and the dataset statistics are detailed in \Cref{tab:example} and \Cref{tab:data}.

\begin{table*}[h]
\begin{center}
\centering
\resizebox{\textwidth}{!}{%
\begin{tabular}{@{\extracolsep{4pt}}lcccccc|c}
\toprule
\textbf{Method}  & \textbf{Model Size} &
\textbf{Filtering} &
\textbf{ROUGE-L} & \textbf{Correctness} & \textbf{Faithfulness} & \textbf{Completeness} & \textbf{Average} \\
\midrule
GPT-3.5 (0-shot) & Unknown&- & 48.5 & 76.1 & 81.0 & 19.4 & 56.3\\
GPT-3.5 (4-shot) & Unknown & \cmark{} & 54.2 & 86.9 & 85.6 & 63.3 & 72.5\\
\midrule
Mixtral-8x7B-Instruct (0-shot)  & 47B& - & 42.6 & 78.0 & 74.1 & 48.3 & 60.8\\
Mixtral-8x7B-Instruct (4-shot) & 47B & \cmark{} & 49.7 & 83.2 & 78.7 & 56.6 & 67.1\\
\midrule
Flan-T5-XL (FT) & 3B &\xmark{}& 56.8 & 83.0 & 83.3 & 51.0 & 68.5 \\
Flan-T5-XL (FT) & 3B &\cmark{} & 58.9 (+2.1) & 85.3 (+2.3) & 85.6 (+2.3) & 64.4 (+13.4) & 73.6 (+5.0) \\
\quad + \ours{} (Ours) & 3B & \cmark{} &\textbf{59.9 (+3.1)} & \textbf{87.6 (+4.6)} & \textbf{87.8 (+4.5)} & \textbf{71.6 (+20.6)} & \textbf{76.7 (+8.2)} \\
\midrule
Mistral-7B-Instruct (FT) & 7B& \xmark{}  & 51.4 & 89.7 & 82.2 & 78.4 & 75.4 \\
Mistral-7B-Instruct (FT) & 7B& \cmark{}& 52.5 (+1.1) & \textbf{90.5 (+0.8)} & \textbf{85.4 (+3.2)} & 81.2 (+2.8) & 77.4 (+3.0)\\
\quad + \ours{} (Ours) & 7B& \cmark{} & \textbf{53.9 (+2.5)} &	89.6 (-0.1)	& 85.3 (+3.1)& \textbf{82.2 (+3.8)}	& \textbf{77.8 (+3.4)} \\
\bottomrule
\end{tabular}
}
\end{center}
\caption{
Main experiment results for response generation on the \ourdata{} test set across four metrics. \ours{} refers to generating intermediate steps for lists before generating responses.
`FT' refers to fine-tuned models.
Filtering indicates whether model-based filtering was applied to improve the quality of the training set.
}
\label{tab:experiment_result}
\end{table*}

\section{Experiment}

\subsection{Experiment Setup}\label{sec:exp_setup}

We evaluate models of various sizes on \ourdata{}. For the larger LLMs, we use GPT-3.5 \footnote{\href{https://platform.openai.com/docs/models/gpt-3-5-turbo}{gpt-3.5-turbo-0125}} and Mixtral-8x7B-Instruct \footnote{\href{https://huggingface.co/mistralai/Mixtral-8x7B-Instruct-v0.1}{Mixtral-8x7B-Instruct-v0.1}}\citep{jiang2024mixtral} in the 0- and 4-shot setting, where 4-shot examples are randomly selected from samples with four different list types.
For the smaller LLMs, we fine-tune Flan-T5-XL \citep{Chung2022ScalingIL} and Mistral-7B-Instruct\footnote{\href{https://huggingface.co/mistralai/Mistral-7B-Instruct-v0.2}{Mistral-7B-Instruct-v0.2}} \citep{Jiang2023Mistral7} on the \ourdata{} training set.
We adopt QLoRA \citep{dettmers2024qlora} with a learning rate of 5e-4 for Flan-T5-XL over 10 epochs and 2e-5 for Mistral-7B-Instruct over 2 epochs. 
For retrieval-augmented generation, we apply LlamaIndex \citep{Liu_LlamaIndex_2022} with `all-mpnet-base-v2' \citep{Reimers2019SentenceBERTSE} as the passage retriever. 
We set the top-$K$ to 3, achieving a recall@3 of 93.0\% on our training set. Questions without relevant passages retrieved are considered as `unanswerable'.
An example input and output with intermediate steps are shown in \Cref{tab:prompt_example}.

\subsection{Evaluation Metric}

We evaluate responses generated by models on the \ourdata{} test set using both non-LLM and LLM-based evaluation. 
For non-LLM evaluation, we select ROUGE-L \citep{lin2004rouge}, which measures the lexical overlap between reference and generated responses.
Additionally, recent work \citep{liu2023gpteval,kim2023prometheus} shows that advanced LLMs can perform fine-grained evaluations, such as detecting hallucinations or missing information, aligning well with human judgments.
Therefore, we adopt the LLM-based evaluation using GPT-4 \citep{OpenAI2023GPT4TR} to measure whether models generate correct responses (`correctness'), whether the responses are solely based on the relevant passage (`faithfulness'), and whether the responses include all the necessary information (`completeness'). Details are in \Cref{app:prompt_for_response_evaluation}.

\subsection{Experimental Results}
We present evaluation results in Table \ref{tab:experiment_result}. 
Notably, fine-tuned language models significantly outperform larger language models. 
For instance, the performance of Mixtral-8x7B-Instruct with 4-shot examples lags behind fine-tuned Flan-T5-XL and Mistral-7B-Instruct by approximately 10.0\% in average score.
This underscores the importance of fine-tuning models to deepen the understanding of nuanced semantic relations in list information for generating responses. 
Our findings further confirm the ability of fine-tuned efficient language models to outperform larger ones on specific tasks, consistent with \citet{li2024small,fu2024tiny}.

The results also demonstrate that model-based filtering of the training data consistently results in performance improvements across two models and four metrics, despite using almost half the number of training samples.
Specifically, Flan-T5-XL trained on the filtered dataset outperforms the baseline by up to 5.0\% on average. Notably, filtering particularly helps to reduce incomplete response generation, achieving a 13.4\% improvement.

Additionally, our \ours{} method, which generates intermediate steps for list information, helps further improve performance. 
For instance, Flan-T5-XL with \ours{} achieves a 3.2\% higher performance than without \ours{} across four metrics on average.
Although the correctness and faithfulness of Mistral-7B-Instruct slightly decrease after applying \ours{}, its overall performance still improves by 0.4\%.

Moreover, models based on Flan-T5-XL achieve higher ROUGE-L (59.9\% vs. 53.9\%) and faithfulness (87.8\% vs. 85.3\%) scores compared to those trained on Mistral-7B-Instruct.
This could be partially because Mistral-7B-Instruct is more verbose than Flan-T5-XL, often producing unnecessary phrases. 
Conversely, this verbosity of Mistral-7B-Instruct models rather helps produce more correct (87.6\% vs. 89.6\%) and complete (71.6\% vs. 82.2\%) responses than Flan-T5-XL, highlighting a trade-off between base language model choices.

\section{Analysis}

\subsection{Performance on Different List Types}

\begin{figure*}[h]
\centering
\subfloat[ROUGE-L]{\includegraphics[width=.49\textwidth]{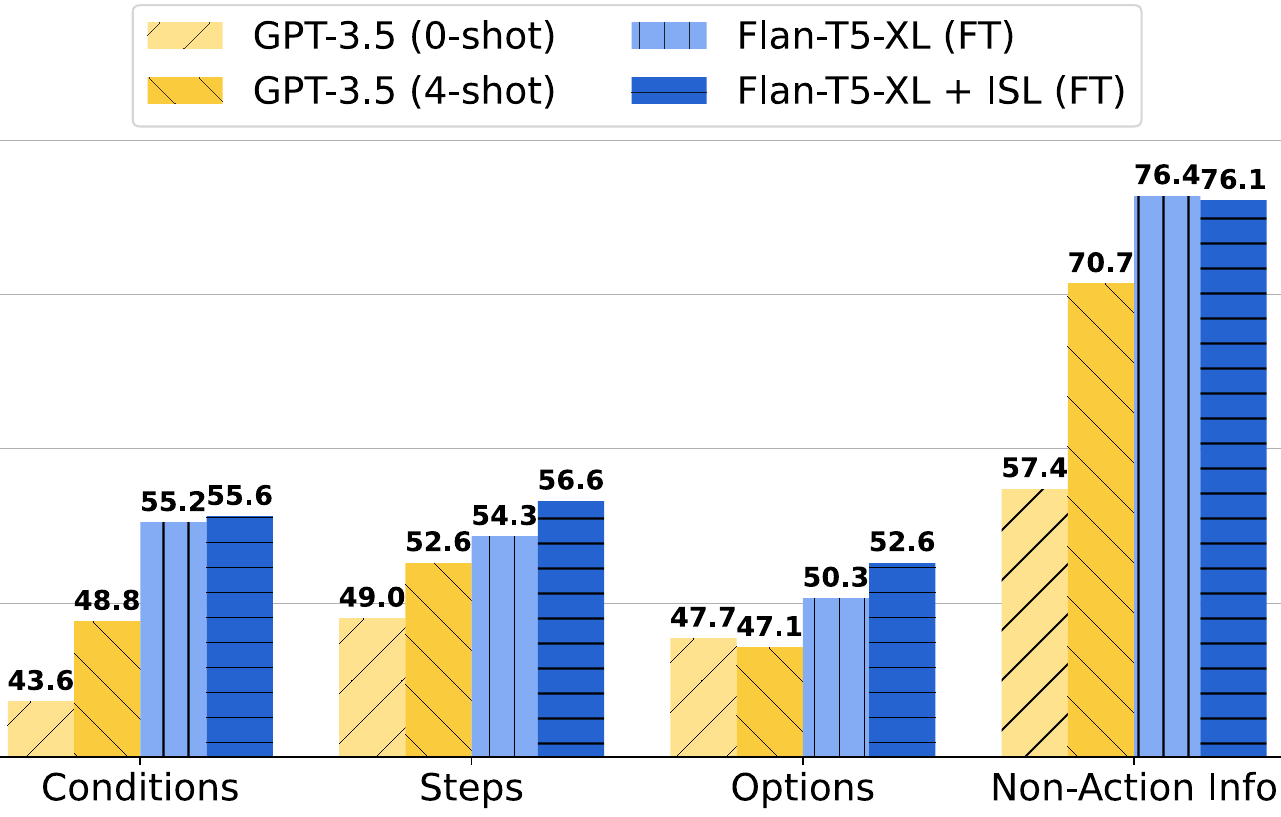}}
\hspace{.01\textwidth}
\subfloat[Correctness]
{\includegraphics[width=.49\textwidth]{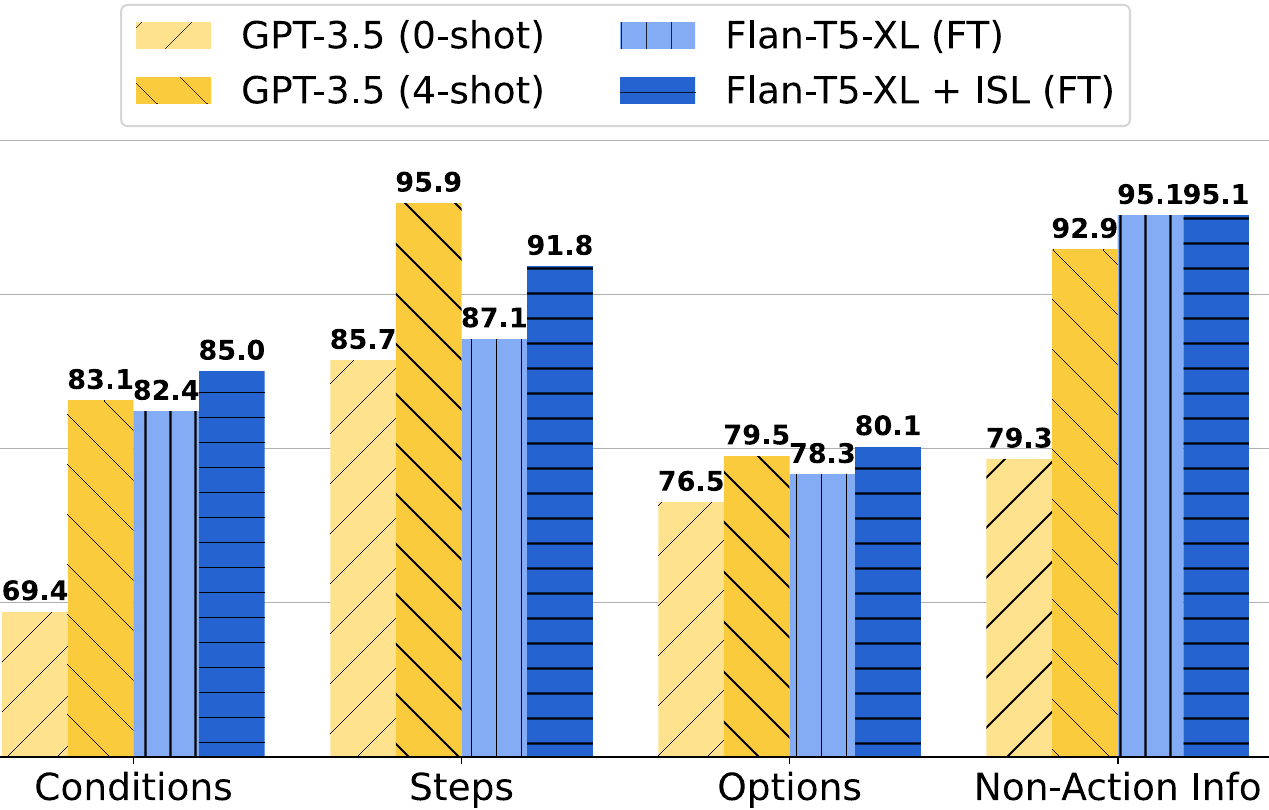}}
\hfill
\subfloat[Faithfulness]{\includegraphics[width=.49\textwidth]{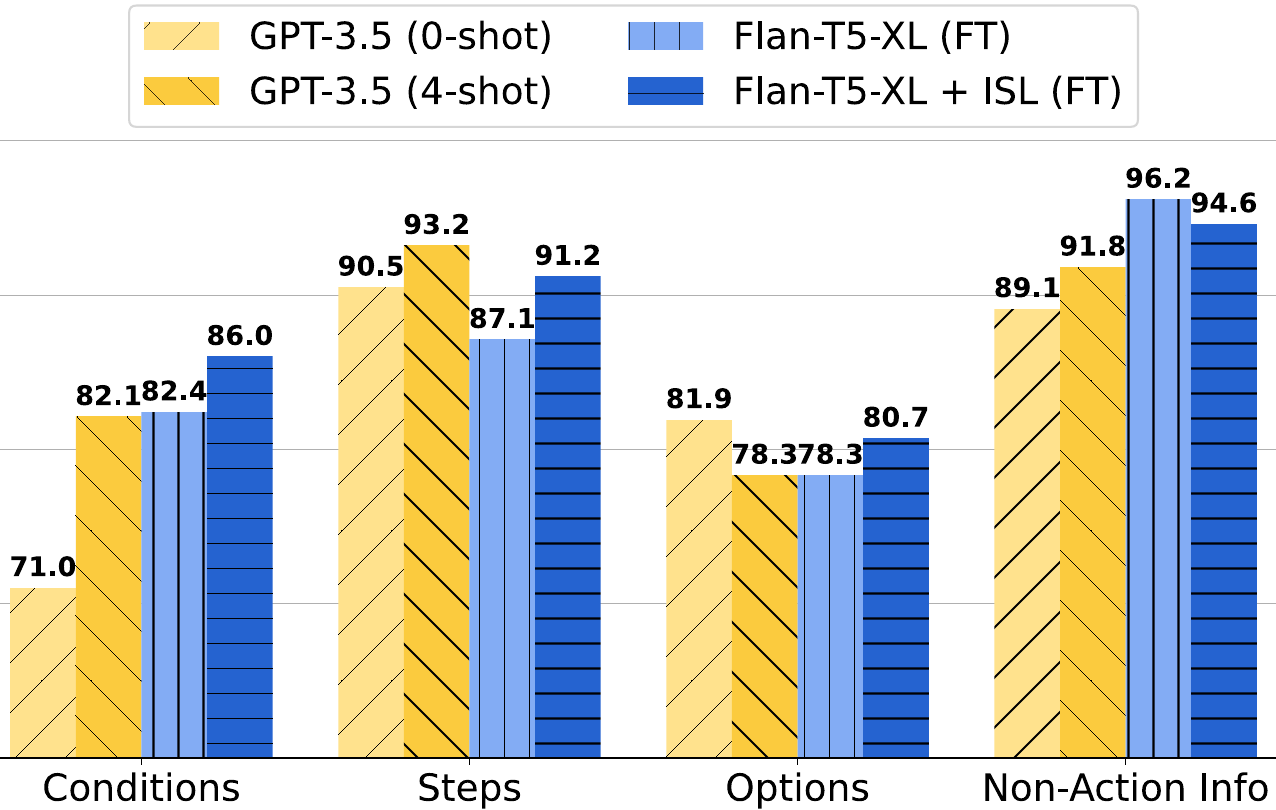}}
\hspace{.01\textwidth}
\subfloat[Completeness]{\includegraphics[width=.49\textwidth]{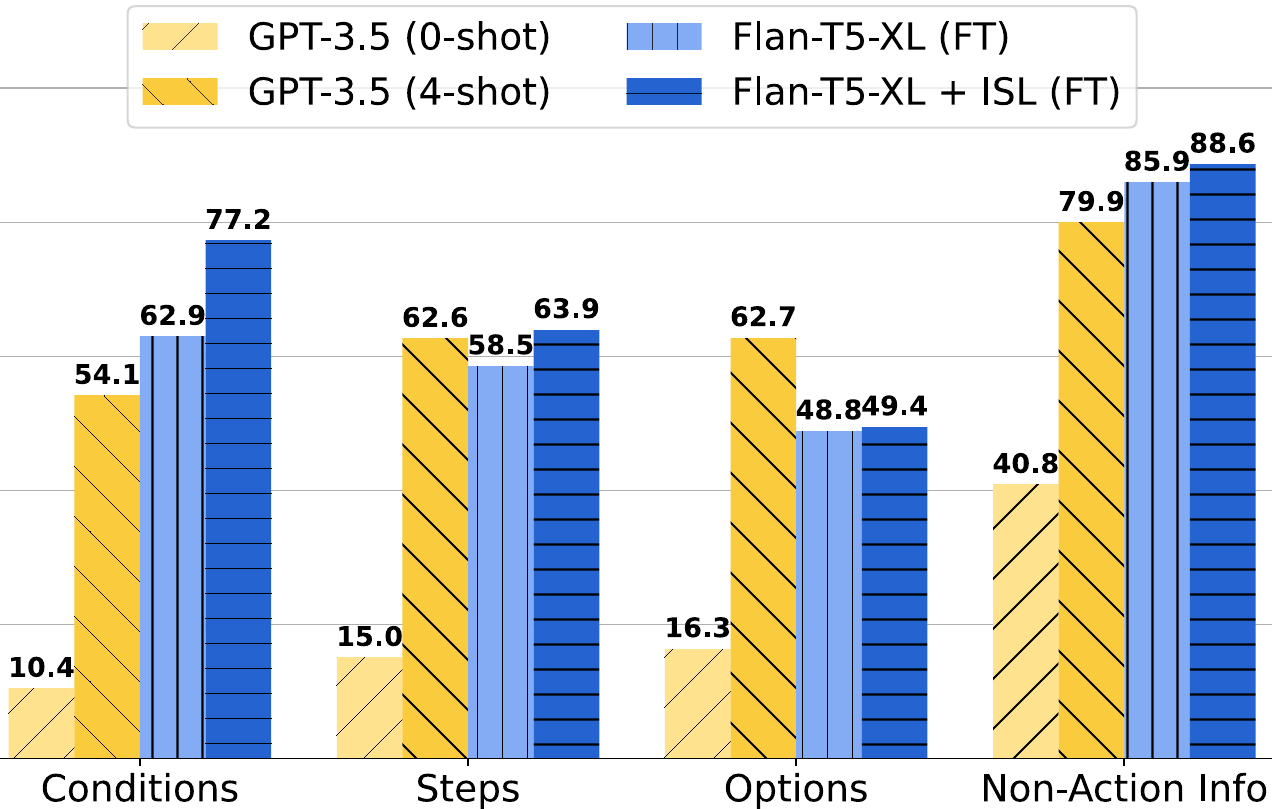}}
\caption{
Performance breakdown across the different list types.
}
\label{fig:performance_list_type}
\end{figure*}

Different list types necessitate distinct styles of user questions and corresponding intermediate steps. 
To understand the performance disparities, we analyzed four list types from the \ourdata{} test set: conditions, steps, options, and non-action information. \Cref{fig:performance_list_type} illustrates that GPT-3.5 particularly struggles with responding over condition and non-action information lists compared to fine-tuned Flan-T5-XL by a large margin.

We observe that leveraging ISL consistently and significantly improves performance, with two exceptions in the non-action information list type regarding ROUGE-L and faithfulness.
Specifically, Flan-T5-XL with ISL outperforms the baseline by up to 2.6\% in correctness, 3.6\% in faithfulness, and 14.3\% in completeness on condition lists, highlighting the benefit of generating intermediate steps for more accurate responses. However, the exceptions observed in non-action information lists suggest that these lists do not typically require complex intermediate steps, such as tracking user-to-item status or logical relations. As a result, training the model on these lists might lead to the generation of unnecessary information in responses, thereby decreasing performance.

\begin{table}[t]
\begin{center}
\centering
\resizebox{1\columnwidth}{!}{%
\begin{tabular}{@{\extracolsep{4pt}}lcccc|c}
\toprule
\textbf{Method} & \textbf{R-L} & \textbf{Correct} & \textbf{Faithful} & \textbf{Complete} & \textbf{Avg} \\
\midrule
\multicolumn{6}{c}{\textit{Seen Domain}} \\
\midrule
GPT-3.5 (0-shot) & 50.4 & 76.5 & 82.2  & 20.5 & 57.4\\
\midrule
Flan-T5-XL (FT) & 60.8 & 87.4 & 87.6 & 68.6 & 76.1 \\
\quad + ISL & \textbf{61.4} & \textbf{89.8} & \textbf{89.8} & \textbf{75.9} & \textbf{79.2} \\
\midrule
Mistral-7B-Ins (FT) & 52.6 & 90.1 & 83.1 & 83.8 & 77.4 \\
\quad + ISL & \textbf{54.5} & \textbf{90.3} & \textbf{84.9} & \textbf{86.2} & \textbf{79.0} \\
\midrule
\multicolumn{6}{c}{\textit{Unseen Domain}} \\
\midrule
GPT-3.5 (0-shot) & 48.2 & 75.6 & 79.5 & 18.0 & 55.3\\
\midrule
Flan-T5-XL (FT) & 56.4 & 82.8 & 83.1 & 59.3 & 70.4 \\
\quad + ISL & \textbf{58.0} & \textbf{84.8} & \textbf{85.3} & \textbf{66.5} & \textbf{73.6} \\
\midrule
Mistral-7B-Ins (FT) & 52.5 & \textbf{91.1} & \textbf{88.4} & \textbf{78.1} & \textbf{77.5} \\
\quad + ISL & \textbf{53.2} & 88.6 & 85.9 & 77.3 & 76.3 \\
\bottomrule
\end{tabular}
}
\end{center}
\caption{
Performance comparison on seen/unseen data.
}
\label{tab:seen_unseen}
\end{table}

\begin{table*}[t]
\begin{center}
\centering
\resizebox{\textwidth}{!}{%
\begin{tabular}{@{\extracolsep{4pt}}ll}
\toprule
\textbf{Gold Passages} & \textbf{Evaluation on Generated Responses} \\
\midrule
\begin{tabular}[c]{@{}l@{}} \textbf{Title:} Running a limited company\\ Other changes you must report\\ You must tell HMRC if:\\  $\bullet{}$ your business' contact details change - \\ for example, your name, gender, business name\\ $\bullet{}$ you appoint an accountant or tax adviser\end{tabular} & \begin{tabular}[c]{@{}l@{}}\textbf{Question:} I have recently changed the name of my business. \\Do I need to inform HMRC about this update?\\ \\ \textbf{Generated Response:} Yes, you need to inform HMRC about the \\ name change of your business \textcolor{red}{within 30 days or you could face} \\ \textcolor{red}{a financial penalty.}\\ \\ \textbf{Evaluation:} Correct, \textcolor{red}{Unfaithful}, Complete\end{tabular} \\
\midrule
\begin{tabular}[c]{@{}l@{}} \textbf{Title:} How to Correct or Update Your FAFSA Application\\ How do I make changes to my FAFSA form?\\ Submit your new information:\\ $\bullet{}$ Write in the corrections or updates on your paper SAR, \\ sign it , and mail it to the address provided on the SAR .\\ \textcolor{red}{$\bullet{}$ Check with the financial aid office at the school you }\\  \textcolor{red}{plan to attend; the school might be able to make changes} \\  \textcolor{red}{for you electronically.}\end{tabular} & \begin{tabular}[c]{@{}l@{}}\textbf{Question:} I need to make some corrections and updates to my FAFSA \\ form. What should I do next with my paper SAR?\\ \\ \textbf{Generated Response:} Next, you should write in the corrections or \\ updates on your paper SAR, sign it, and mail it to the address provided \\ on the SAR.\\ \\ \textbf{Evaluation:} Correct, Faithful, 
\textcolor{red}{Incomplete}\end{tabular}  \\
\bottomrule
\end{tabular}
}
\end{center}
\caption{
Sample errors, e.g., `unfaithful' or `incomplete', in generated responses by fine-tuned Flan-T5-XL with ISL. Text in \textcolor{red}{red} marks the information related to the errors.
}
\label{tab:error_cases}
\end{table*}

\subsection{Performance on Seen and Unseen Domains}

To evaluate the generalizability of our approach on unseen domains, we use samples from the ConditionalQA corpus for the seen domain and samples from the MultiDoc2Dial corpus for the unseen domain.
As shown in \Cref{tab:seen_unseen}, the fine-tuned models with ISL consistently outperform those without ISL on seen data, achieving up to a 3.1\% higher average score.
This trend extends to Flan-T5-XL with ISL on the unseen domain, showing a 3.2\% increase, which demonstrates the robustness of fine-tuning Flan-T5-XL with ISL.
However, Mistral-7B-Instruct with ISL struggles on the unseen domain, with a 0.8\% decrease in predicting correct intermediate steps, highlighting the need for further improvements in generalizability.

\subsection{Error Cases}

Analyzing error cases provides valuable insights into the limitations of our models. In \Cref{tab:error_cases}, fine-tuned Flan-T5-XL with ISL generates the phrase `within 30 days or you could face a financial penalty', which is not present in the gold passage. Additionally, the model omits crucial information such as `check with the financial aid office ...', which should have been included in the responses. Future work could explore more sophisticated approaches, such as preference optimization \citep{wu2024fine, rafailov2024direct}, to improve the generation of more faithful and complete responses grounding over list information.

\section{Related Work}

\paragraph{Document-Grounded Dialogue}

Our work is generally related to the document-grounded dialogue task \citep{Feng2020Doc2DialAG, Feng2021MultiDoc2DialMD, wu2021dialki, gao2022unigdd, zhao2023causal, Le2023ImprovedIO}. While previous research has largely concentrated on dialogue systems that respond to information-seeking user questions based on plain text knowledge, they often overlook user requests that involve verifying conditions found in support documents. Although some work, such as ShARC \citep{Saeidi2018InterpretationON} and ConditionalQA \citep{Sun2021ConditionalQAAC}, has begun to address this by focusing on conditional content presented as lists within documents, these tasks are still somewhat distant from real-world scenarios involving differentiating from other types of text and structured content. Our work bridges this gap by recognizing a broader range of list types and nuanced semantic relationships indicated by lists. We propose a novel approach that leverages large language models (LLMs) to better handle these complexities, thereby supporting further research in LLM-based dialogue systems.

\paragraph{Intermediate Steps}

Our approach involving \Ours{} (\ours{}) for QA systems is generally related to generating intermediate reasoning for large language models 
\citep{Wei2022ChainOT, kojima2022large, zhang2022automatic, zhou2022least, yao2022react, huang-chang-2023-towards, yu-etal-2023-alert, wang2023learning}, 
which enhances the reasoning ability of large language models.
While most previous works focus on using intermediate reasoning in free-form text, structured approaches to intermediate reasoning have been proposed for mainly for specific tasks such as code generation \citep{Li2023StructuredCP}.
Our work specifically focuses on understanding nuanced semantic relations for QA tasks based on list information.
Additionally, our setup is within the context of data augmentation, featuring a development data simulation pipeline.
We emphasize a pipelined approach integrating data augmentation and efficient fine-tuning to enhance the performance of smaller LLMs, particularly in handling list semantics.

\section{Conclusion}

We present an novel pipeline-based approach to enhance question answering systems by addressing the nuanced challenges posed by list-based content. Our primary contributions include the introduction of the \ourdata{} dataset, a novel benchmark designed to assess QA systems' ability to effectively handle and respond to list information. Additionally, we develop the \Ours{} (\ours{}) method, which mirrors human interpretive processes for list items. Our experiments show that our approach, based on efficient fine-tuned models, consistently outperforms baseline approaches. By emphasizing the importance of QA systems' ability to handle list-based content with dynamic and nuanced semantics, our work paves ways for future research to further refine QA systems and expand their applicability across various domains.

\section*{Limitations}
There are certain limitations in current scope of this work: 
(1) Although we currently handle only two types of logical relations in conditional lists, namely `and' and `or', there are more diverse logical relation types, such as `nor' or nested relations, in passages containing lists. We plan to investigate these in future work.
(2) While we focus only on single-turn QA tasks in this paper, multi-turn dialogues grounding over structured lists, where systems need to respond considering dialogue history, can be more practical. We leave this exploration of multi-turn dialogues grounding over structured lists for future work.
(3) Evaluating models' generation across `correctness', `faithfulness', and `completeness' using GPT-4 is costly \citep{tang2024minicheck}, which hinders more extensive evaluations, and is somewhat less accurate compared to human evaluation. In the future, we aim to develop an automatic evaluation method, which is less expensive and more accurate, for structured list-grounded question answering systems.

\section*{Ethical Considerations}
The dataset and models presented in this work have some ethical considerations: (1) The data simulation process should ensure diversity and avoid representation biases by incorporating input from humans with diverse backgrounds; (2) The question answering systems should provide transparent explanations for its responses to build appropriate trust with users; (3) Further testing is needed to proactively evaluate fairness and safety issues before deployment to real users, in order to prevent harm. 

\bibliography{anthology}

\begin{thebibliography}{45}
\providecommand{\natexlab}[1]{#1}

\bibitem[{Campos et~al.(2020)Campos, Otegi, Soroa, Deriu, Cieliebak, and Agirre}]{campos2020doqa}
Jon~Ander Campos, Arantxa Otegi, Aitor Soroa, Jan Deriu, Mark Cieliebak, and Eneko Agirre. 2020.
\newblock \href {https://doi.org/10.18653/v1/2020.acl-main.652} {{D}o{QA} - accessing domain-specific {FAQ}s via conversational {QA}}.
\newblock In \emph{ACL}.

\bibitem[{Chen et~al.(2021)Chen, Chen, Smiley, Shah, Borova, Langdon, Moussa, Beane, Huang, Routledge, and Wang}]{chen2021finqa}
Zhiyu Chen, Wenhu Chen, Charese Smiley, Sameena Shah, Iana Borova, Dylan Langdon, Reema Moussa, Matt Beane, Ting-Hao Huang, Bryan Routledge, and William~Yang Wang. 2021.
\newblock \href {https://doi.org/10.18653/v1/2021.emnlp-main.300} {{F}in{QA}: A dataset of numerical reasoning over financial data}.
\newblock In \emph{EMNLP}.

\bibitem[{Choi et~al.(2024)Choi, Lee, Jin, and Kim}]{choi2024gpts}
Juhwan Choi, Eunju Lee, Kyohoon Jin, and YoungBin Kim. 2024.
\newblock \href {https://aclanthology.org/2024.findings-eacl.2} {{GPT}s are multilingual annotators for sequence generation tasks}.
\newblock In \emph{Findings of EACL}.

\bibitem[{Chung et~al.(2024)Chung, Hou, Longpre, Zoph, Tay, Fedus, Li, Wang, Dehghani, Brahma, Webson, Gu, Dai, Suzgun, Chen, Chowdhery, Castro-Ros, Pellat, Robinson, Valter, Narang, Mishra, Yu, Zhao, Huang, Dai, Yu, Petrov, Chi, Dean, Devlin, Roberts, Zhou, Le, and Wei}]{Chung2022ScalingIL}
Hyung~Won Chung, Le~Hou, Shayne Longpre, Barret Zoph, Yi~Tay, William Fedus, Yunxuan Li, Xuezhi Wang, Mostafa Dehghani, Siddhartha Brahma, Albert Webson, Shixiang~Shane Gu, Zhuyun Dai, Mirac Suzgun, Xinyun Chen, Aakanksha Chowdhery, Alex Castro-Ros, Marie Pellat, Kevin Robinson, Dasha Valter, Sharan Narang, Gaurav Mishra, Adams Yu, Vincent Zhao, Yanping Huang, Andrew Dai, Hongkun Yu, Slav Petrov, Ed~H. Chi, Jeff Dean, Jacob Devlin, Adam Roberts, Denny Zhou, Quoc~V. Le, and Jason Wei. 2024.
\newblock \href {http://jmlr.org/papers/v25/23-0870.html} {Scaling instruction-finetuned language models}.
\newblock \emph{JMLR}.

\bibitem[{Dettmers et~al.()Dettmers, Pagnoni, Holtzman, and Zettlemoyer}]{dettmers2024qlora}
Tim Dettmers, Artidoro Pagnoni, Ari Holtzman, and Luke Zettlemoyer.
\newblock \href {https://proceedings.neurips.cc/paper_files/paper/2023/file/1feb87871436031bdc0f2beaa62a049b-Paper-Conference.pdf} {Qlora: Efficient finetuning of quantized llms}.
\newblock In \emph{NeurIPS}.

\bibitem[{Feng et~al.(2021)Feng, Patel, Wan, and Joshi}]{Feng2021MultiDoc2DialMD}
Song Feng, Siva~Sankalp Patel, Hui Wan, and Sachindra Joshi. 2021.
\newblock \href {https://doi.org/10.18653/v1/2021.emnlp-main.498} {{M}ulti{D}oc2{D}ial: Modeling dialogues grounded in multiple documents}.
\newblock In \emph{EMNLP}.

\bibitem[{Feng et~al.(2020)Feng, Wan, Gunasekara, Patel, Joshi, and Lastras}]{Feng2020Doc2DialAG}
Song Feng, Hui Wan, Chulaka Gunasekara, Siva Patel, Sachindra Joshi, and Luis Lastras. 2020.
\newblock \href {https://doi.org/10.18653/v1/2020.emnlp-main.652} {doc2dial: A goal-oriented document-grounded dialogue dataset}.
\newblock In \emph{EMNLP}.

\bibitem[{Fu et~al.(2024)Fu, Laskar, Khasanova, Chen, and Tn}]{fu2024tiny}
Xue-Yong Fu, Md~Tahmid~Rahman Laskar, Elena Khasanova, Cheng Chen, and Shashi Tn. 2024.
\newblock \href {https://doi.org/10.18653/v1/2024.naacl-industry.33} {Tiny titans: Can smaller large language models punch above their weight in the real world for meeting summarization?}
\newblock In \emph{NAACL}.

\bibitem[{Gao et~al.(2022)Gao, Zhang, and Lam}]{gao2022unigdd}
Chang Gao, Wenxuan Zhang, and Wai Lam. 2022.
\newblock \href {https://doi.org/10.18653/v1/2022.acl-short.66} {Unigdd: A unified generative framework for goal-oriented document-grounded dialogue}.
\newblock In \emph{ACL}.

\bibitem[{Guo et~al.(2023)Guo, Jin, Liu, Huang, Shi, Yu, Liu, Li, Xiong, Xiong et~al.}]{guo2023evaluating}
Zishan Guo, Renren Jin, Chuang Liu, Yufei Huang, Dan Shi, Linhao Yu, Yan Liu, Jiaxuan Li, Bojian Xiong, Deyi Xiong, et~al. 2023.
\newblock \href {https://doi.org/10.48550/arXiv.2310.19736} {Evaluating large language models: A comprehensive survey}.
\newblock \emph{ArXiv}.

\bibitem[{He et~al.(2024)He, Lin, Gong, Jin, Zhang, Lin, Jiao, Yiu, Duan, Chen et~al.}]{he2023annollm}
Xingwei He, Zhenghao Lin, Yeyun Gong, Alex Jin, Hang Zhang, Chen Lin, Jian Jiao, Siu~Ming Yiu, Nan Duan, Weizhu Chen, et~al. 2024.
\newblock \href {https://doi.org/10.18653/v1/2024.naacl-industry.15} {Annollm: Making large language models to be better crowdsourced annotators}.
\newblock In \emph{NAACL}.

\bibitem[{Huang and Chang(2023)}]{huang-chang-2023-towards}
Jie Huang and Kevin Chen-Chuan Chang. 2023.
\newblock \href {https://doi.org/10.18653/v1/2023.findings-acl.67} {Towards reasoning in large language models: A survey}.
\newblock In \emph{Findings of ACL}.

\bibitem[{Jiang et~al.(2024)Jiang, Sablayrolles, Roux, Mensch, Savary, Bamford, Chaplot, Casas, Hanna, Bressand et~al.}]{jiang2024mixtral}
Albert~Q Jiang, Alexandre Sablayrolles, Antoine Roux, Arthur Mensch, Blanche Savary, Chris Bamford, Devendra~Singh Chaplot, Diego de~las Casas, Emma~Bou Hanna, Florian Bressand, et~al. 2024.
\newblock \href {https://doi.org/10.48550/arXiv.2401.04088} {Mixtral of experts}.
\newblock \emph{ArXiv}.

\bibitem[{Jiang et~al.(2023)Jiang, Sablayrolles, Mensch, Bamford, Chaplot, de~Las~Casas, Bressand, Lengyel, Lample, Saulnier, Lavaud, Lachaux, Stock, Scao, Lavril, Wang, Lacroix, and Sayed}]{Jiang2023Mistral7}
Albert~Qiaochu Jiang, Alexandre Sablayrolles, Arthur Mensch, Chris Bamford, Devendra~Singh Chaplot, Diego de~Las~Casas, Florian Bressand, Gianna Lengyel, Guillaume Lample, Lucile Saulnier, L'elio~Renard Lavaud, Marie-Anne Lachaux, Pierre Stock, Teven~Le Scao, Thibaut Lavril, Thomas Wang, Timoth{\'e}e Lacroix, and William~El Sayed. 2023.
\newblock \href {https://doi.org/10.48550/arXiv.2310.06825} {Mistral 7b}.
\newblock \emph{ArXiv}.

\bibitem[{Kim et~al.(2024)Kim, Shin, Cho, Jang, Longpre, Lee, Yun, Shin, Kim, Thorne et~al.}]{kim2023prometheus}
Seungone Kim, Jamin Shin, Yejin Cho, Joel Jang, Shayne Longpre, Hwaran Lee, Sangdoo Yun, Seongjin Shin, Sungdong Kim, James Thorne, et~al. 2024.
\newblock \href {https://doi.org/10.48550/arXiv.2310.08491} {Prometheus: Inducing evaluation capability in language models}.
\newblock In \emph{ICLR}.

\bibitem[{Kojima et~al.(2022)Kojima, Gu, Reid, Matsuo, and Iwasawa}]{kojima2022large}
Takeshi Kojima, Shixiang~Shane Gu, Machel Reid, Yutaka Matsuo, and Yusuke Iwasawa. 2022.
\newblock \href {https://proceedings.neurips.cc/paper_files/paper/2022/file/8bb0d291acd4acf06ef112099c16f326-Paper-Conference.pdf} {Large language models are zero-shot reasoners}.
\newblock \emph{NeurIPS}.

\bibitem[{Le et~al.(2023)Le, Guo, Xu, and Ritter}]{Le2023ImprovedIO}
Duong~Minh Le, Ruohao Guo, Wei Xu, and Alan Ritter. 2023.
\newblock \href {https://doi.org/10.18653/v1/2023.acl-long.561} {Improved instruction ordering in recipe-grounded conversation}.
\newblock In \emph{ACL}.

\bibitem[{Lewis et~al.(2020)Lewis, Perez, Piktus, Petroni, Karpukhin, Goyal, K{\"u}ttler, Lewis, Yih, Rockt{\"a}schel et~al.}]{lewis2020retrieval}
Patrick Lewis, Ethan Perez, Aleksandra Piktus, Fabio Petroni, Vladimir Karpukhin, Naman Goyal, Heinrich K{\"u}ttler, Mike Lewis, Wen-tau Yih, Tim Rockt{\"a}schel, et~al. 2020.
\newblock \href {https://proceedings.neurips.cc/paper_files/paper/2020/file/6b493230205f780e1bc26945df7481e5-Paper.pdf} {Retrieval-augmented generation for knowledge-intensive nlp tasks}.
\newblock \emph{NeurIPS}.

\bibitem[{Li et~al.(2024)Li, Zhang, Bubeck, Pathuri, and Menache}]{li2024small}
Beibin Li, Yi~Zhang, S{\'e}bastien Bubeck, Jeevan Pathuri, and Ishai Menache. 2024.
\newblock \href {https://doi.org/10.48550/arXiv.2405.20347} {Small language models for application interactions: A case study}.
\newblock \emph{ArXiv}.

\bibitem[{Li et~al.(2023)Li, Li, Li, and Jin}]{Li2023StructuredCP}
Jia Li, Ge~Li, Yongming Li, and Zhi Jin. 2023.
\newblock \href {https://doi.org/10.48550/arXiv.2305.06599} {Structured chain-of-thought prompting for code generation}.
\newblock \emph{ArXiv}.

\bibitem[{Lin(2004)}]{lin2004rouge}
Chin-Yew Lin. 2004.
\newblock \href {https://aclanthology.org/W04-1013} {Rouge: A package for automatic evaluation of summaries}.
\newblock In \emph{Text summarization branches out}.

\bibitem[{Liu et~al.(2023{\natexlab{a}})Liu, Ning, Teng, Liu, Zhou, and Zhang}]{liu2023evaluating}
Hanmeng Liu, Ruoxi Ning, Zhiyang Teng, Jian Liu, Qiji Zhou, and Yue Zhang. 2023{\natexlab{a}}.
\newblock \href {https://doi.org/10.48550/arXiv.2304.03439} {Evaluating the logical reasoning ability of chatgpt and gpt-4}.
\newblock \emph{ArXiv}.

\bibitem[{Liu(2022)}]{Liu_LlamaIndex_2022}
Jerry Liu. 2022.
\newblock Llamaindex.
\newblock \url{https://github.com/jerryjliu/llama_index}.

\bibitem[{Liu et~al.(2023{\natexlab{b}})Liu, Iter, Xu, Wang, Xu, and Zhu}]{liu2023gpteval}
Yang Liu, Dan Iter, Yichong Xu, Shuohang Wang, Ruochen Xu, and Chenguang Zhu. 2023{\natexlab{b}}.
\newblock \href {https://doi.org/10.18653/v1/2023.emnlp-main.153} {Gpteval: Nlg evaluation using gpt-4 with better human alignment}.
\newblock In \emph{EMNLP}.

\bibitem[{McCarthy and Jarvis(2010)}]{mccarthy2010mtld}
Philip~M McCarthy and Scott Jarvis. 2010.
\newblock \href {https://link.springer.com/article/10.3758/BRM.42.2.381} {Mtld, vocd-d, and hd-d: A validation study of sophisticated approaches to lexical diversity assessment}.
\newblock \emph{Behavior research methods}.

\bibitem[{Nan et~al.(2022)Nan, Hsieh, Mao, Lin, Verma, Zhang, Kry{\'s}ci{\'n}ski, Schoelkopf, Kong, Tang et~al.}]{nan2022fetaqa}
Linyong Nan, Chiachun Hsieh, Ziming Mao, Xi~Victoria Lin, Neha Verma, Rui Zhang, Wojciech Kry{\'s}ci{\'n}ski, Hailey Schoelkopf, Riley Kong, Xiangru Tang, et~al. 2022.
\newblock \href {https://doi.org/10.1162/tacl_a_00446} {Fetaqa: Free-form table question answering}.
\newblock \emph{TACL}.

\bibitem[{Oh et~al.(2024)Oh, Shin, Ko, Lee, and Seo}]{oh2023ktrl+}
Hanseok Oh, Haebin Shin, Miyoung Ko, Hyunji Lee, and Minjoon Seo. 2024.
\newblock \href {https://doi.org/10.18653/v1/2024.naacl-long.134} {Ktrl+ f: Knowledge-augmented in-document search}.
\newblock In \emph{NAACL}.

\bibitem[{OpenAI(2022)}]{chatgpt}
OpenAI. 2022.
\newblock Chatgpt blog post.
\newblock \url{https://openai.com/blog/chatgpt}.

\bibitem[{OpenAI(2023)}]{OpenAI2023GPT4TR}
OpenAI. 2023.
\newblock \href {https://doi.org/10.48550/arXiv.2303.08774} {Gpt-4 technical report}.
\newblock \emph{ArXiv}.

\bibitem[{Os{\'e}s~Grijalba et~al.(2024)Os{\'e}s~Grijalba, Ure{\~n}a-L{\'o}pez, Mart{\'\i}nez~C{\'a}mara, and Camacho-Collados}]{oses-grijalba-etal-2024-question}
Jorge Os{\'e}s~Grijalba, L.~Alfonso Ure{\~n}a-L{\'o}pez, Eugenio Mart{\'\i}nez~C{\'a}mara, and Jose Camacho-Collados. 2024.
\newblock \href {https://aclanthology.org/2024.lrec-main.1179} {Question answering over tabular data with {D}ata{B}ench: A large-scale empirical evaluation of {LLM}s}.
\newblock In \emph{LREC-COLING}.

\bibitem[{Qin et~al.(2023)Qin, Zhang, Zhang, Chen, Yasunaga, and Yang}]{qin2023chatgpt}
Chengwei Qin, Aston Zhang, Zhuosheng Zhang, Jiaao Chen, Michihiro Yasunaga, and Diyi Yang. 2023.
\newblock \href {https://doi.org/10.18653/v1/2023.emnlp-main.85} {Is chatgpt a general-purpose natural language processing task solver?}
\newblock In \emph{EMNLP}.

\bibitem[{Rafailov et~al.(2024)Rafailov, Sharma, Mitchell, Manning, Ermon, and Finn}]{rafailov2024direct}
Rafael Rafailov, Archit Sharma, Eric Mitchell, Christopher~D Manning, Stefano Ermon, and Chelsea Finn. 2024.
\newblock \href {https://doi.org/10.48550/arXiv.2305.18290} {Direct preference optimization: Your language model is secretly a reward model}.
\newblock In \emph{NeurIPS}.

\bibitem[{Reimers and Gurevych(2019)}]{Reimers2019SentenceBERTSE}
Nils Reimers and Iryna Gurevych. 2019.
\newblock \href {https://doi.org/10.18653/v1/D19-1410} {Sentence-bert: Sentence embeddings using siamese bert-networks}.
\newblock In \emph{EMNLP}.

\bibitem[{Saeidi et~al.(2018)Saeidi, Bartolo, Lewis, Singh, Rockt{\"a}schel, Sheldon, Bouchard, and Riedel}]{Saeidi2018InterpretationON}
Marzieh Saeidi, Max Bartolo, Patrick Lewis, Sameer Singh, Tim Rockt{\"a}schel, Mike Sheldon, Guillaume Bouchard, and Sebastian Riedel. 2018.
\newblock \href {https://doi.org/10.18653/v1/D18-1233} {Interpretation of natural language rules in conversational machine reading}.
\newblock In \emph{EMNLP}.

\bibitem[{Sun et~al.(2021)Sun, Cohen, and Salakhutdinov}]{Sun2021ConditionalQAAC}
Haitian Sun, William~W. Cohen, and Ruslan Salakhutdinov. 2021.
\newblock \href {https://doi.org/10.18653/v1/2022.acl-long.253} {Conditionalqa: A complex reading comprehension dataset with conditional answers}.
\newblock In \emph{ACL}.

\bibitem[{Tang et~al.(2024)Tang, Laban, and Durrett}]{tang2024minicheck}
Liyan Tang, Philippe Laban, and Greg Durrett. 2024.
\newblock \href {https://doi.org/10.48550/arXiv.2404.10774} {Minicheck: Efficient fact-checking of llms on grounding documents}.
\newblock \emph{ArXiv}.

\bibitem[{Wang and Lu(2023)}]{wang2023learning}
Tianduo Wang and Wei Lu. 2023.
\newblock \href {https://doi.org/10.18653/v1/2023.acl-short.106} {Learning multi-step reasoning by solving arithmetic tasks}.
\newblock In \emph{ACL}.

\bibitem[{Wei et~al.(2022)Wei, Wang, Schuurmans, Bosma, hsin Chi, Xia, Le, and Zhou}]{Wei2022ChainOT}
Jason Wei, Xuezhi Wang, Dale Schuurmans, Maarten Bosma, Ed~Huai hsin Chi, F.~Xia, Quoc Le, and Denny Zhou. 2022.
\newblock \href {https://dl.acm.org/doi/10.5555/3600270.3602070} {Chain of thought prompting elicits reasoning in large language models}.
\newblock In \emph{NeurIPS}.

\bibitem[{Wu et~al.(2024)Wu, Hu, Shi, Dziri, Suhr, Ammanabrolu, Smith, Ostendorf, and Hajishirzi}]{wu2024fine}
Zeqiu Wu, Yushi Hu, Weijia Shi, Nouha Dziri, Alane Suhr, Prithviraj Ammanabrolu, Noah~A Smith, Mari Ostendorf, and Hannaneh Hajishirzi. 2024.
\newblock \href {https://papers.neurips.cc/paper_files/paper/2023/file/b8c90b65739ae8417e61eadb521f63d5-Paper-Conference.pdf} {Fine-grained human feedback gives better rewards for language model training}.
\newblock In \emph{NeurIPS}.

\bibitem[{Wu et~al.(2021)Wu, Lu, Hajishirzi, and Ostendorf}]{wu2021dialki}
Zeqiu Wu, Bo-Ru Lu, Hannaneh Hajishirzi, and Mari Ostendorf. 2021.
\newblock \href {https://doi.org/10.18653/v1/2021.emnlp-main.140} {Dialki: Knowledge identification in conversational systems through dialogue-document contextualization}.
\newblock In \emph{EMNLP}.

\bibitem[{Yao et~al.(2023)Yao, Zhao, Yu, Du, Shafran, Narasimhan, and Cao}]{yao2022react}
Shunyu Yao, Jeffrey Zhao, Dian Yu, Nan Du, Izhak Shafran, Karthik Narasimhan, and Yuan Cao. 2023.
\newblock \href {https://doi.org/10.48550/arXiv.2210.03629} {React: Synergizing reasoning and acting in language models}.
\newblock In \emph{ICLR}.

\bibitem[{Yu et~al.(2023)Yu, Wang, Golovneva, AlKhamissi, Verma, Jin, Ghosh, Diab, and Celikyilmaz}]{yu-etal-2023-alert}
Ping Yu, Tianlu Wang, Olga Golovneva, Badr AlKhamissi, Siddharth Verma, Zhijing Jin, Gargi Ghosh, Mona Diab, and Asli Celikyilmaz. 2023.
\newblock \href {https://doi.org/10.18653/v1/2023.acl-long.60} {{ALERT}: Adapt language models to reasoning tasks}.
\newblock In \emph{ACL}.

\bibitem[{Zhang et~al.(2022)Zhang, Zhang, Li, and Smola}]{zhang2022automatic}
Zhuosheng Zhang, Aston Zhang, Mu~Li, and Alex Smola. 2022.
\newblock \href {https://doi.org/10.48550/arXiv.2210.03493} {Automatic chain of thought prompting in large language models}.
\newblock \emph{ArXiv}.

\bibitem[{Zhao et~al.(2023)Zhao, Yu, Yu, Li, Li, Wang, Huang, Li, and Zhang}]{zhao2023causal}
Yingxiu Zhao, Bowen Yu, Haiyang Yu, Bowen Li, Jinyang Li, Chao Wang, Fei Huang, Yongbin Li, and Nevin~L Zhang. 2023.
\newblock \href {https://doi.org/10.18653/v1/2023.emnlp-main.443} {Causal document-grounded dialogue pre-training}.
\newblock In \emph{EMNLP}.

\bibitem[{Zhou et~al.(2023)Zhou, Sch{\"a}rli, Hou, Wei, Scales, Wang, Schuurmans, Bousquet, Le, and Chi}]{zhou2022least}
Denny Zhou, Nathanael Sch{\"a}rli, Le~Hou, Jason Wei, Nathan Scales, Xuezhi Wang, Dale Schuurmans, Olivier Bousquet, Quoc Le, and Ed~Chi. 2023.
\newblock \href {https://doi.org/10.48550/arXiv.2205.10625} {Least-to-most prompting enables complex reasoning in large language models}.
\newblock \emph{ICLR}.

\end{thebibliography}
\newpage

\appendix

\section{Details of Logical Relation Classifier}\label{abs:logical_relation_classifier}

We found that classifying logical relations between list items is surprisingly difficult for large language models using in-context learning. \Cref{tab:logical_relation_classifier} shows that Mistral-7B-Instruct, and even the larger Mixtral-8x7B-Instruct, with 8-shot in-context examples, struggle with the seemingly simple task of classifying `And' or `Or,' achieving F1 scores lower than 30 on the validation samples. To address this issue, we fine-tuned Flan-T5-XL using 72 manually annotated training samples and achieved an F1 score of 78.0 on 32 manually curated validation samples.

\begin{table}[h]
\begin{center}
\centering
\resizebox{1\columnwidth}{!}{%
\begin{tabular}{@{\extracolsep{4pt}}lccc}
\toprule
\textbf{Model} & \textbf{And F1} & \textbf{Or F1} & \textbf{Avg F1} \\
\midrule
Flan-T5-XL (72-shot FT) & \textbf{77.6} &  \textbf{78.4} & \textbf{78.0}\\
Mistral-7B-Instruct (8-shot IC) & 34.9 & 22.7 & 28.8 \\
Mixtral-8x7B-Instruct (8-shot IC) & 49.4 & 4.1 & 26.8 \\
\bottomrule
\end{tabular}
}
\end{center}
\caption{
Comparison of models for classifying logical relations on 32 manually curated validation samples. `FT' refers to fine-tuning, and `IC' refers to in-context learning.
}
\label{tab:logical_relation_classifier}
\end{table}

\section{Prompt for Model-based Filtering}\label{app:prompt_for_data_verification}
\Cref{tab:prompt_for_data_verification} describes the prompt used for filtering out noisy samples in which user questions are considered unanswerable or the system responses are found to be incorrect, unfaithful, or incomplete on the given context.

\begin{table*}[t]
\begin{center}
\centering
\resizebox{\textwidth}{!}{%
\begin{tabular}{@{\extracolsep{4pt}}p{16cm}}
\toprule
You will be evaluating a system's response to a user question, given some context. Here is the context:\\\\
<context>\\
\{\{CONTEXT\}\}\\
</context>\\\\

Here is the user's question:\\\\
<question>\\
\{\{QUESTION\}\}\\
</question>\\\\

And here is the system's response:\\\\
<response>\\
\{\{RESPONSE\}\}\\
</response>\\\\
\textbf{First, determine if the question can be answered based solely on the information provided in the context.} Output your reasoning inside <answerability\_reasoning>. Then output "answerable" or "unanswerable" inside <answerable> tags. \\\\

\textbf{Next, if the question is answerable, evaluate the system's response across three dimensions:}\\
- Correctness: Is the response factually correct based on the context?\\
- Faithfulness: Does the response avoid claiming anything not directly supported by the context?\\
- Completeness: Does the response include all relevant information from the context to fully answer the question? \\
If the question is unanswerable, output "NA" for each of the three dimensions. For each dimension, first output your reasoning inside  <correctness\_reasoning>, <faithfulness\_reasoning> and <completeness\_reasoning> tags. 
Then output your assessment (correct/incorrect/NA, faithful/unfaithful/NA, complete/incomplete/NA) inside <correctness>, <faithfulness> and <completeness> tags. \\
\bottomrule
\end{tabular}
}
\end{center}
\caption{
Prompt for model-based filtering.
This involves checking whether questions are answerable, and system responses are correct, faithful, and complete.
}
\label{tab:prompt_for_data_verification}
\end{table*}

\section{Prompt for Response Evaluation}\label{app:prompt_for_response_evaluation}
\Cref{tab:prompt_for_response_evaluation} describes the prompt evaluating whether generated responses are correct, faithful, or complete based on the given context and the user question. 

\begin{table*}[t]
\begin{center}
\centering
\resizebox{\textwidth}{!}{%
\begin{tabular}{@{\extracolsep{4pt}}p{16cm}}
\toprule
You will be evaluating a system's response to a user question, given some context. Here is the context:\\\\
<context>\\
\{\{CONTEXT\}\}\\
</context>\\\\

Here is the user's question:\\\\
<question>\\
\{\{QUESTION\}\}\\
</question>\\\\

And here is the system's response:\\\\
<response>\\
\{\{RESPONSE\}\}\\
</response>\\\\
\textbf{Evaluate the system's response across three dimensions:}\\
- Correctness: Is the response factually correct based on the context?\\
- Faithfulness: Does the response avoid claiming anything not directly supported by the context?\\
- Completeness: Does the response include all relevant information from the context to fully answer the question? \\If the question is unanswerable, output "NA" for each of the three dimensions. For each dimension, first output your reasoning inside  <correctness\_reasoning>, <faithfulness\_reasoning> and <completeness\_reasoning> tags. Then output your assessment (correct/incorrect/NA, faithful/unfaithful/NA, complete/incomplete/NA) inside <correctness>, <faithfulness> and <completeness> tags. \\
\bottomrule
\end{tabular}
}
\end{center}
\caption{
Prompt for response evaluation.
This involves checking whether system responses are correct, faithful, and complete.
}
\label{tab:prompt_for_response_evaluation}
\end{table*}

\end{document}